\documentclass[10pt, a4paper]{article}
\usepackage{lrec}
\usepackage{multibib}
\newcites{languageresource}{Language Resources}
\usepackage{graphicx}
\usepackage{tabularx}
\usepackage{soul}

\usepackage{epstopdf}
\usepackage[utf8]{inputenc}

\usepackage{hyperref}
\usepackage{xstring}

\usepackage{color}

\usepackage{CJKutf8}
\usepackage{graphicx}
\usepackage{subcaption}
\usepackage{booktabs}
\usepackage{multirow}

\newcommand{\Tstrut}{\rule{0pt}{2.3ex}}
\newcommand{\Ustrut}{\rule{0pt}{3.8ex}}
\newcommand{\cc}[1]{\begin{CJK*}{UTF8}{bkai}\hbox{#1}\end{CJK*}}
\newcommand{\cncpt}[2]{#1$\vert$#2}

\title{CA-EHN: Commonsense Analogy from E-HowNet}

\name{Peng-Hsuan Li, Tsan-Yu Yang, Wei-Yun Ma}

\address{
Academia Sinica \\
jacobvsdanniel@iis.sinica.edu.tw, s0920331239@gmail.com, ma@iis.sinica.edu.tw \\
}

\abstract{
Embedding commonsense knowledge is crucial for end-to-end models to generalize inference beyond training corpora. However, existing word analogy datasets have tended to be handcrafted, involving permutations of hundreds of words with only dozens of pre-defined relations, mostly morphological relations and named entities. In this work, we model commonsense knowledge down to word-level analogical reasoning by leveraging E-HowNet, an ontology that annotates 88K Chinese words with their structured sense definitions and English translations. We present CA-EHN, the first commonsense word analogy dataset containing 90,505 analogies covering 5,656 words and 763 relations. Experiments show that CA-EHN stands out as a great indicator of how well word representations embed commonsense knowledge. The dataset is publicly available at \url{https://github.com/ckiplab/CA-EHN}. \\
\newline 
}

\begin{document}

\maketitleabstract

\section{Introduction}

Commonsense reasoning is fundamental for natural language agents to generalize inference beyond training corpora. Although the natural language inference (NLI) task \citelanguageresource{Bowman:EMNLP2015,Williams:NAACL2018} has proved a good pre-training objective for sentence representations \cite{Conneau:EMNLP2017}, commonsense coverage is implicit and limited by the amount of annotated sentence pairs. Furthermore, most models are still end-to-end, relying heavily on word representations to provide background world knowledge.

Therefore, it is desirable to model commonsense knowledge down to word-level analogical reasoning. In this sense, existing analogy benchmarks are lackluster. For Chinese analogy (CA), the simplified Chinese dataset CA8 \citelanguageresource{Li:ACL2018} and the traditional Chinese dataset CA-Google \citelanguageresource{Chen:LREC2018}, translated from English \citelanguageresource{Mikolov:arxiv2013}, contain only a few dozen relations, most of which are either morphological, e.g., a shared prefix, or about named entities, e.g., capital-country.

However, commonsense knowledge bases such as WordNet \citelanguageresource{Miller:ACM1995} and ConceptNet \citelanguageresource{Speer:LREC2012} have long annotated relations in our lexicon. Among them, E-HowNet \citelanguageresource{Chen:ONTOLEX2005,Ma:LREC2018}, extended from HowNet \citelanguageresource{Dong:NLPKE2003}, currently annotates 88K traditional Chinese words with their structured definitions and English translations.

In this paper, we propose an algorithm to extract accurate analogies from E-HowNet with refinements from linguists. We present CA-EHN, the first commonsense analogy dataset containing 90,505 analogies covering 5,656 words and 763 relations. In the experiments, we show that it is useful to embed more commonsense knowledge and that CA-EHN tests this aspect of word embedding.

\section{Related Work}

In this work, we use word sense definitions from the E-HowNet \citelanguageresource{Chen:ONTOLEX2005,Ma:LREC2018} ontology as well as further linguist refinements to construct our commonsense word analogy corpus. Compared to the WordNet \citelanguageresource{Miller:ACM1995} gloss, E-HowNet has structured definitions for word senses, each of which can be parsed into a definition graph. These graphs are fundamentally different from that of ConceptNet \citelanguageresource{Speer:LREC2012}. In ConceptNet, there is one huge graph where each node is a concept (words) and each edge is a relation induced by two concepts. For example, there is a \textbf{capable-of} edge from \textbf{bird} to \textbf{fly}. In this work, for each word sense, we create its defining graph, where edges represent modifying attributes. For example, there is a \textbf{predication} edge from \textbf{animal} to \textbf{fly} in the defining graph of \textbf{bird}. More detailed and precise descriptions are given in Section \ref{sec:ehownet} and Section \ref{sec:method}.

Notable Chinese word analogy datasets include CA8 \citelanguageresource{Li:ACL2018} and CA-Google \citelanguageresource{Chen:LREC2018}. The former is created by Chinese annotators, and the later is translated from English analogies to Chinese. Both of their analogies are essentially the permutation of word pairs that span only a few dozens of relations, mostly regarding named entities and morphology. In contrast, the analogies of CA-EHN are extracted from the E-HowNet lexical ontology and span hundreds of common sense relations. Table \ref{tab:benchmark} shows detailed statistics of these word analogy corpora.

\section{E-HowNet}
\label{sec:ehownet}

E-HowNet 2.0\footnote{\cc{廣義知網2.0版}(\url{http://ehownet.iis.sinica.edu.tw})} \citelanguageresource{Ma:LREC2018} consists of two major parts: A lexicon of \textit{words}, \textit{concepts}, and \textit{attributes} (Table~\ref{tab:lexicon}), and a taxonomy of concepts (Figure~\ref{fig:taxonomy}).

\begin{table*}
\centering
\begin{tabular}{@{}lll@{}}
\toprule
\Tstrut Token & Type & Definitions \\
\midrule
\Tstrut telic                             & attribute & \\
\Ustrut \cc{協}                           & word      & \#1:\{\cncpt{help}{\cc{幫助}}\} \\
\Tstrut                                   &           & \#2:\{\cncpt{community}{\cc{團體}}\} \\
\Ustrut \cncpt{\cc{駿馬}}{ExcellentSteed} & concept   & \{\cncpt{\cc{馬}}{{horse}}:qualification=\{\cncpt{HighQuality}{\cc{優質}}\}\} \\
\Ustrut \cc{實驗室}                       & word      & \#1:\{\cncpt{InstitutePlace}{\cc{場所}}:telic=\{or(\{\cncpt{experiment}{\cc{實驗}}: \\
                                          &           & ~~~~~~~location=\{\textasciitilde{}\}\},\{\cncpt{research}{\cc{研究}}:location=\{\textasciitilde{}\}\})\}\} \\
\bottomrule
\end{tabular}
\caption{\label{tab:lexicon} E-HowNet lexicon.}
\end{table*}

\subsection{Lexicon}
\label{sec:lexicon}

The E-HowNet lexicon consists of 88K words, 4K concepts, and dozens of attributes. These three types of tokens can be readily distinguished by token names: Words, such as \cc{人} and \cc{雞}, are entirely in Chinese. Concepts, such as \cncpt{human}{\cc{人}} and \cncpt{\cc{雞}}{chicken}, contain a vertical bar and an English string in their name. (The order of English and Chinese does not matter in this work.) Attributes, such as telic and theme, are always in English.

In the lexicon, each word is annotated with one or more word senses, and each word sense has a structured \textit{definition}. Each definition consists of concepts connected by attributes. Furthermore, every concept also comes with one such structured definition. In essence, words are defined by concepts, and concepts are recursively defined by more abstract concepts. Table~\ref{tab:lexicon} shows a part of the lexicon, with gradually more complex definition examples:
\begin{itemize}
\item The attribute telic has no definition.
\item The word \cc{協} has two senses. The first (help) is trivially defined by \cncpt{help}{\cc{幫助}}, and the second (association) by \cncpt{community}{\cc{團體}}.
\item The concept \cncpt{\cc{駿馬}}{ExcellentSteed} is defined as \cncpt{\cc{馬}}{horse} modified by \cncpt{HighQuality}{\cc{優質}} with the qualification attribute.
\item The word \cc{實驗室} has only one sense (laboratory), defined as an \cncpt{InstitutePlace}{\cc{場所}} used as the location for \cncpt{experiment}{\cc{實驗}} or \cncpt{research}{\cc{研究}}.
\end{itemize}

In this work, we use E-HowNet word sense definitions to extract commonsense analogies (Section~\ref{sec:method}). Besides, word senses are annotated with their English translations, which could be used to transfer our extracted analogies to English multi-word expressions.

\subsection{Taxonomy}
\label{sec:taxonomy}

\begin{figure}[!ht]
    \centering
    \includegraphics[width=0.95\columnwidth]{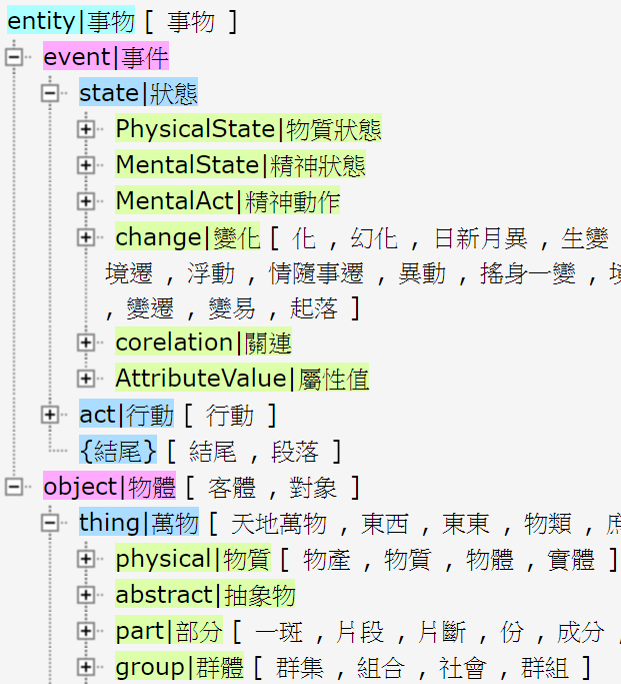}
	 \caption{E-HowNet taxonomy.}
    \label{fig:taxonomy}
\end{figure}

The E-HowNet taxonomy organizes concepts into a tree. Figure~\ref{fig:taxonomy} shows the partially expanded taxonomy. The words beside each node have senses defined trivially by that concept. For example, one definition of \cc{東西} is simply \{\cncpt{thing}{\cc{萬物}}\}.

In the experiments, we infuse E-HowNet taxonomy to distributed word representations and analyze performance changes across word analogy benchmarks (Section~\ref{sec:commonsense_benchmarking}).

\begin{figure*}
    \centering
    \includegraphics[width=1.75\columnwidth]{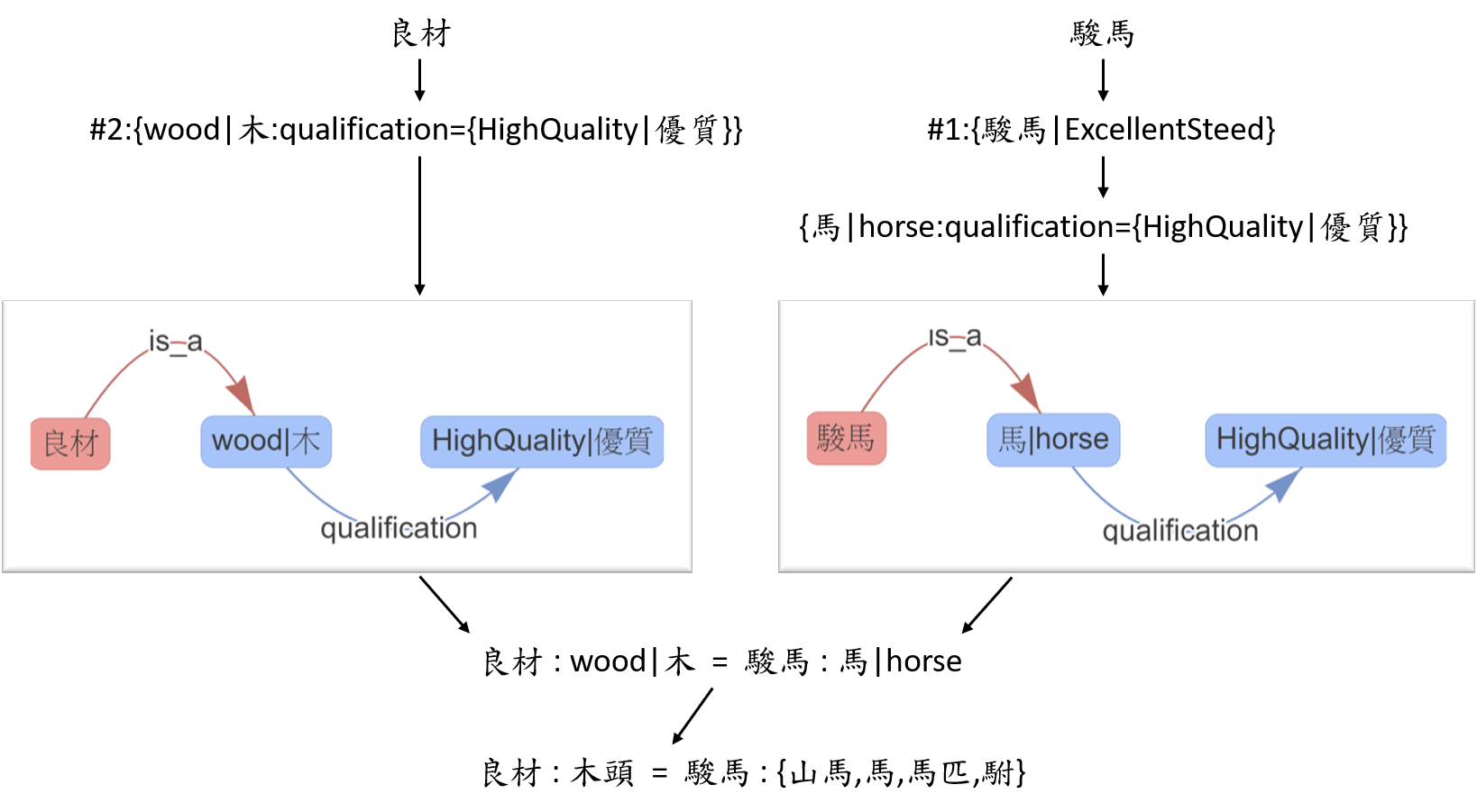}
	 \caption{Commonsense analogy extraction.}
    \label{fig:process}
\end{figure*}

\begin{figure*}
    \centering
    \begin{subfigure}{1.2\columnwidth}
        \centering
        \includegraphics[width=\columnwidth]{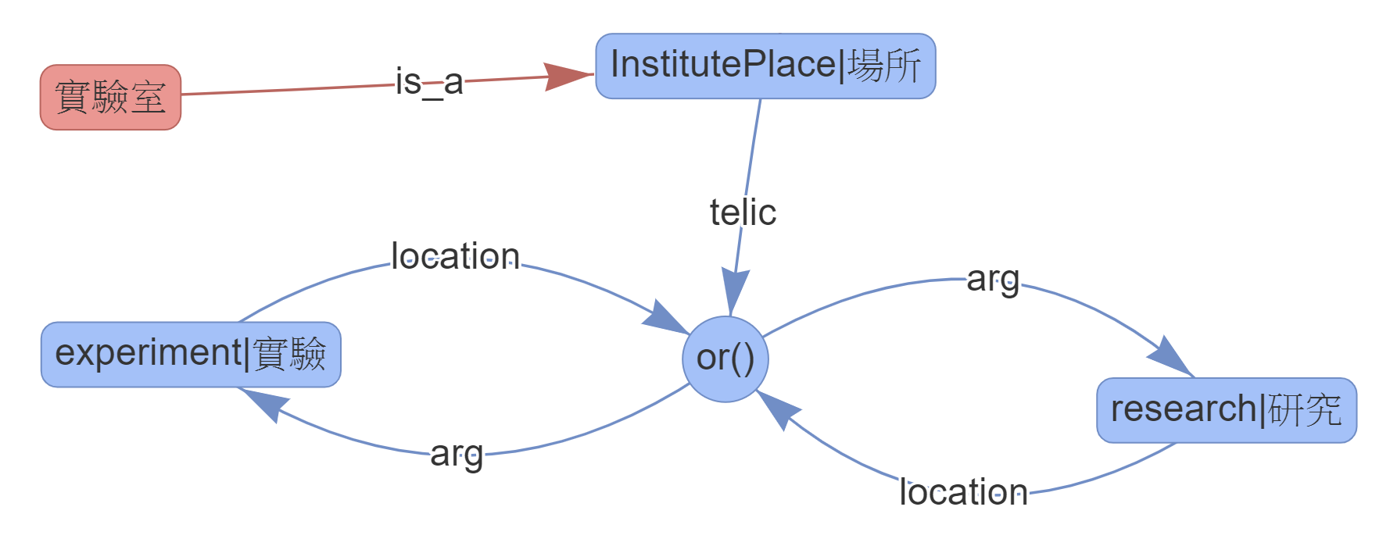}
        \caption{\label{fig:parse:laboratory}\cc{實驗室} (laboratory).}
    \end{subfigure}
    \hspace{.01\columnwidth}
    \begin{subfigure}{.35\columnwidth}
        \centering
        \includegraphics[width=\columnwidth]{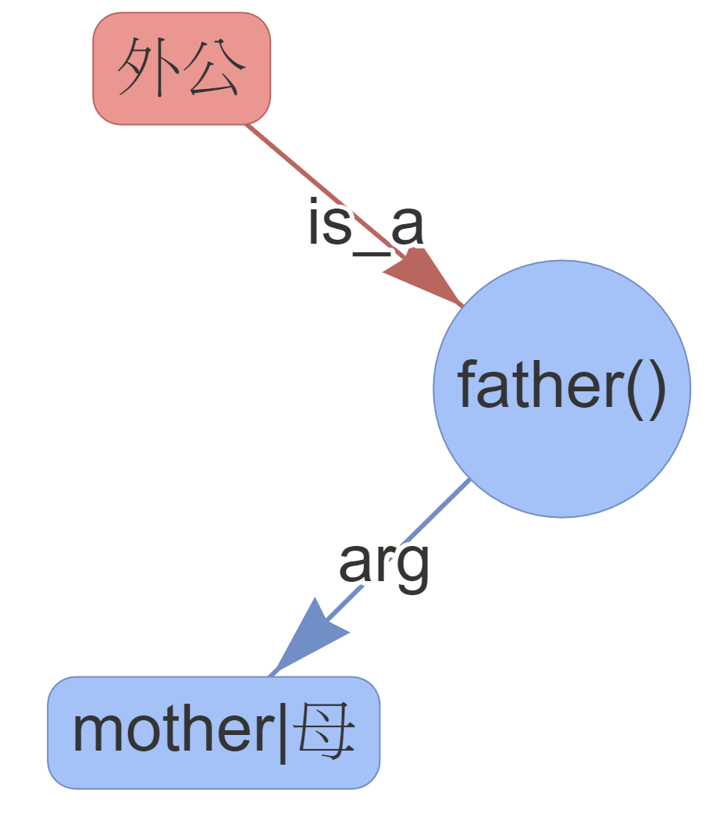}
        \caption{\label{fig:parse:grandpa}\cc{外公}\\ (maternal grandpa).}
    \end{subfigure}
    \caption{Sample parsed definition graphs.}
    \label{fig:definition}
\end{figure*}

\section{Commonsense Analogy}
\label{sec:method}

We extract word analogies with rich coverage of words and commonsense relations by comparing word sense definitions (Section~\ref{sec:lexicon}). The extraction algorithm is further refined with multiple filters and linguist annotations.

\subsection{Analogy Extraction}

Before refinement, for each sense pair of two words, we try to extract analogies with the following five steps. This process is illustrated in Figure~\ref{fig:process}.

\paragraph{Definition Expansion} A definition is expanded if it contains only one concept. For example, \cc{駿馬} is defined simply as \{\cncpt{\cc{駿馬}}{ExcellentSteed}\}. Such trivial definitions would only lead to trivial analogies equating synonym pairs. We resolve this problem by replacing the definitions of those words with the definitions of their defining concepts. For example, the definition of \cc{駿馬} is replaced by \{\cncpt{\cc{馬}}{horse}:qualification=\{\cncpt{HighQuality}{\cc{優質}}\}\}, i.e., the definition of \cncpt{\cc{駿馬}}{ExcellentSteed}.

\paragraph{Definition Parsing} Each definition is parsed into a directed graph. Each node in the graph is either a word, a concept, or a \textit{function}, e.g., or() at the bottom of Table~\ref{tab:lexicon}. Each edge either links to an attribute modifier, e.g., :telic=, or connects a function node with its argument nodes. Figure~\ref{fig:definition} shows some more parsed definition graphs.

\paragraph{Graph Comparison} The two definition graphs are compared to see if they differ only in one concept node. If they do, the two (word, concept) pairs are analogical to one another. For example, since the graph of \cc{良材} sense\#2 (the good timber) and the expanded graph of \cc{駿馬} sense\#1 (an excellent steed) differ only in \cncpt{wood}{\cc{木}} and \cncpt{\cc{馬}}{horse}, we extract the following \textit{concept analogy} -- \cc{良材}:\cncpt{wood}{\cc{木}}=\cc{駿馬}:\cncpt{\cc{馬}}{horse}.

\paragraph{Left Expansion} The left concept in the concept analogy is expanded into synonym words, i.e., words that have one sense defined trivially by it. For example, there is only one word \cc{木頭} defined as \{\cncpt{wood}{\cc{木}}\}. Thus after expansion, there is still only one analogy: \cc{良材}:\cc{木頭}=\cc{駿馬}:\cncpt{\cc{馬}}{horse}. Most of the time, this step yields multiple analogies per concept analogy.

\paragraph{Right Expansion} Finally, the right concept in each analogy is also expanded into synonym words. However, this time we do not use them to form multiple analogies. Instead, the word list is kept as a \textit{synset}. For example, as \cc{山馬}, \cc{馬}, \cc{馬匹}, \cc{駙} all have one sense defined as \{\cncpt{\cc{馬}}{horse}\}, the final analogy becomes \cc{良材}:\cc{木頭}=\cc{駿馬}:\{\cc{山馬},\cc{馬},\cc{馬匹},\cc{駙}\}. The reason why not making multiple analogies in this final step is explained in Section~\ref{sec:evaluation}.

\subsection{Embedding Evaluation}
\label{sec:evaluation}

Word analogies are typically used for the intrinsic evaluation of word embeddings. For each analogy $w_1$:$w_2$=$w_3$:$w_4$, the tuple ($w_1$, $w_2$, $w_3$) is called an analogy question and $w_4$ is the answer. A word embedding must predict the answer as the word of which vector is closest to $v_3+v_2-v_1$.

In extracting word analogies from E-HowNet, the left expansion step creates plausible analogy questions, but the above embedding evaluation will not work if the right expansion step creates multiple analogies with the same analogy question. This is why the final step keeps the expanded words in a synset. When evaluating embeddings on our benchmark, a predicted word is considered correct as long as it belongs to the synset.

\begin{table*}
\centering
\begin{tabular}{@{}llll@{}}
\toprule
\Tstrut \cc{滴答}          & \cc{時鐘} & \cc{鼕鼕}          & \{\cc{鼓}\}\\
\Tstrut (tick-tock)        & (clock)   & (rat-tat)          & (\cncpt{\cc{鼓}}{drum}) \\
\midrule
\Tstrut \cc{聾子}          & \cc{耳}   & \cc{瞎子}          & \{\cc{目}, \cc{眸子}, \cc{眼}, \cc{眼眸}, \cc{眼睛}\} \\
\Tstrut (deaf person)      & (ear)     & (blind person)     & (\cncpt{eye}{\cc{眼}}) \\
\midrule
\Tstrut \cc{外公}          & \cc{母親} & \cc{祖父}          & \{\cc{父}, \cc{父親}, \cc{爸}, \cc{爸爸}, \cc{爹}, \cc{爹爹}, \cc{老子}\} \\
\Tstrut (maternal grandpa) & (mother)  & (paternal grandpa) & (\cncpt{father}{\cc{父}}) \\
\midrule
\Tstrut \cc{蝌蚪}          & \cc{青蛙} & \cc{孑孓}          & \{\cc{斑蚊},\cc{蚊}, \cc{蚊子}, \cc{蚊蟲}\} \\
\Tstrut (tadpole)          & (frog)    & (wriggler)         & (\cncpt{\cc{蚊子}}{mosquito}) \\
\bottomrule
\end{tabular}
\caption{\label{tab:caehn} CA-EHN. (word:word=word:synset)}
\end{table*}

\begin{table*}
\centering
\begin{tabular}{@{}lllrrr@{}}
\toprule
Benchmark & Language & Type & \#analogies & \#words & \#relations \\
\midrule
\multirow{4}{*}{CA8-Morphological} & \multirow{4}{*}{Simplified} &
      reduplication A (morph.)  &     2,554 &    344 &  3 \\
  & & reduplication AB (morph.) &     2,535 &    423 &  3 \\
  & & semi-prefix (morph.)      &     2,553 &    656 & 21 \\
  & & semi-suffix (morph.)      &     2,535 &    727 & 41 \\
\midrule
\multirow{4}{*}{CA8-Semantic} & \multirow{4}{*}{Simplified} &
      geography (entity)        &     3,192 &    305 &  9 \\
  & & history (entity)          &     1,465 &    177 &  4 \\
  & & nature                    &     1,370 &    452 & 10 \\
  & & people (entity)           &     1,609 &    259 &  5 \\
\midrule
\multirow{1}{*}{CA-Google} & \multirow{1}{*}{Traditional*} &
      morph., entity, gender    &    11,126 &    498 & 14 \\
\midrule
CA-EHN & Traditional & commonsense & 90,505 & 5,656 & 763 \\
\bottomrule
\end{tabular}
\caption{\label{tab:benchmark} Analogy benchmarks. *Translated from English.}
\end{table*}

\subsection{Accurate Analogy}

As the core procedure yields an excessively large benchmark, added to the fact that E-HowNet word sense definitions are sometimes inaccurate, we add refinements to the extraction process to extract a feasible sized benchmark of accurate analogies.

\paragraph{Concrete Concepts} At every step of the extraction process, we require every word and concept to be under \cncpt{physical}{\cc{物質}}. As shown in Figure~\ref{fig:taxonomy}, this excludes abstract taxa such as \cncpt{event}{\cc{事件}} and \cncpt{abstract}{\cc{抽象物}}. Thus it filters out words that are often hard to accurately define. This restriction shrinks the benchmark by 50\%.

\paragraph{Common Words} At every step of the extraction process, we require words to occur at least five times in ASBC 4.0 \citelanguageresource{Ma:ROCLING2001}, a segmented traditional Chinese corpus containing 10M words from articles between 1981 and 2007. This eliminates uncommon, ancient words or words with synonymous but uncommon, ancient characters. This restriction further shrinks the remaining benchmark by 90\%.

\paragraph{Concept Analogy Annotation} After introducing the previous two refinements, 36,100 concept analogies are extracted by the graph comparison step. Then, linguists are asked to follow an annotation guideline to label their correctness. 1,000 of these concept analogies are labeled by all four annotators with $\kappa=0.76$, indicating a high inter-annotator agreement. This step results in 25,010 remaining concept analogies.

\paragraph{Synset Annotation} Before concept expansion, every synset needed by the 25,010 concept analogies is checked again to remove words that are not actually synonymous with the defining concept. For example, all words in \{\cc{花草}, \cc{山茶花}, \cc{薰衣草}, \cc{鳶尾花}\} are common words and have a sense defined trivially as \cncpt{FlowerGrass}{\cc{花草}}. However, the last three (camellia, lavender, iris) are judged by the annotator as not synonyms but hyponyms to the concept. So, the synset will be refined to \{\cc{花草}\}. This step also helps eliminate words in a synset that are using their rare senses, as we do not expect embeddings to encode those senses without word sense disambiguation.

\begin{table*}[!t]
\centering
\begin{tabular}{ccccccccc}
\toprule
\multirow{2}{*}{Embedding} & \multicolumn{2}{c}{CA8-Morph.} & \multicolumn{2}{c}{CA8-Semantic} & \multicolumn{2}{c}{CA-Google} & \multicolumn{2}{c}{CA-EHN} \\
 & acc & cov* & acc & cov* & acc & cov* & acc & cov* \\
\midrule
\Tstrut GloVe-Small & 0.085 & 6,917 & 0.296 & 4,274 & 0.381 & 5,367 & 0.033 & 90,505 \\
\Tstrut GloVe-Large & 0.115 & 7,379 & 0.376 & 5,761 & 0.437 & 8,409 & 0.044 & 90,505 \\
\Tstrut SGNS-Large  & 0.178 & 7,379 & 0.374 & 5,761 & 0.502 & 8,409 & 0.051 & 90,505 \\
\bottomrule
\end{tabular}
\caption{\label{tab:embedding_benchmarking} Embedding benchmarking. *The number of analogy questions covered by an embedding.}
\end{table*}

\begin{table*}[!t]
\centering
\begin{tabular}{ccccccccc}
\toprule
\multirow{2}{*}{+E-HowNet} & \multicolumn{2}{c}{CA8-Morph.} & \multicolumn{2}{c}{CA8-Semantic} & \multicolumn{2}{c}{CA-Google} & \multicolumn{2}{c}{CA-EHN} \\
 & acc & $\Delta$acc & acc & $\Delta$acc & acc & $\Delta$acc & acc & $\Delta$acc \\
\midrule
\Tstrut GloVe-Small & 0.113 & +33\% & 0.309 &  +4\% & 0.391 &  +3\% & 0.092 & +179\% \\
\Tstrut GloVe-Large & 0.137 & +19\% & 0.376 &   0\% & 0.418 &  -4\% & 0.113 & +157\% \\
\Tstrut SGNS-Large  & 0.180 &  +1\% & 0.379 &  +1\% & 0.489 &  -3\% & 0.127 & +149\% \\
\bottomrule
\end{tabular}
\caption{\label{tab:ehownet_retrofit_benchmarking} E-HowNet retrofit benchmarking.}
\end{table*}
 
\begin{table*}[!t]
\centering
\begin{tabular}{ccccccccc}
\toprule
\multirow{2}{*}{+HIT-Thesaurus} & \multicolumn{2}{c}{CA8-Morph.} & \multicolumn{2}{c}{CA8-Semantic} & \multicolumn{2}{c}{CA-Google} & \multicolumn{2}{c}{CA-EHN} \\
 & acc & $\Delta$acc & acc & $\Delta$acc & acc & $\Delta$acc & acc & $\Delta$acc \\
\midrule
\Tstrut GloVe-Small & 0.126 & +48\% & 0.340 & +15\% & 0.415 &  +9\% & 0.062 &  +88\% \\
\Tstrut GloVe-Large & 0.150 & +30\% & 0.381 &  +1\% & 0.437 &   0\% & 0.076 &  +73\% \\
\Tstrut SGNS-Large  & 0.204 & +15\% & 0.385 &  +3\% & 0.502 &   0\% & 0.083 &  +63\% \\
\bottomrule
\end{tabular}
\caption{\label{tab:hitthesaurus_retrofit_benchmarking} HIT-Thesaurus retrofit benchmarking.}
\end{table*}

\section{Analyses}

With the proposed extraction algorithm, refinements, and linguists annotations, we collected 90,505 accurate analogies for CA-EHN. Table~\ref{tab:caehn} shows a few samples of the corpus, covering such diverse domains as onomatopoeia, disability, kinship, and zoology. We then compare CA-EHN to existing word analogy datasets shown in Table~\ref{tab:benchmark}.

\subsection{Relation Category}

The relations in the datasets can be classified into three categories:
\begin{itemize}
    \item Morphological relations. \\
    For example, the shared prefix \cc{周} (week): \\
    \cc{一}:\cc{周一}=\cc{二}:\cc{周二}=\cc{三}:\cc{周三}=... \\
    (one : Monday = two : Tuesday = three : Wednesday = ...)
    \item Named entity relations. \\
    For example, states to their currencies: \\
    \cc{美國}:\cc{美元}=\cc{丹麥}:\cc{克朗}=\cc{印度}:\cc{盧比}=... \\
    (US : dollar = Denmark : krone = India : rupee = ...)
    \item Commonsense relations. \\
    For example, the solid-fluid relation: \\
    \cc{冰}:\cc{水}=\cc{雪}:\cc{雨}=\cc{固體}:\cc{液體}=... \\
    (ice : water = snow : rain = solid : fluid = ...)
\end{itemize}

Existing datasets contain mostly morphological (morph.) or named entity (entity) relations. The few exceptions are the nature part of CA8 \citelanguageresource{Li:ACL2018} and the gender part of CA-Google \citelanguageresource{Chen:LREC2018}. In contrast, CA-EHN fully dedicates as an extensive benchmark for commonsense word reasoning.

\subsection{Relation Diversity}

For the total number of covered relations, existing datasets span only dozens of pre-defined relations. Their analogies are then created as the permutations of word pair equations. For example, CA8 uses the province-university relation \\
\begin{itemize}
    \item ~\cc{北京}:\cc{北京大學}=\cc{南京}:\cc{南京大學}=\cc{海南:海南大學}=... \\
    (Beijin : Peking University = Nanjing : Nanjing University = Hainan : Hainan University = ...)
\end{itemize}
to create more than two hundred analogies.

\begin{figure*}[!t]
    \centering
    \begin{subfigure}{.44\columnwidth}
        \centering
        \includegraphics[width=\columnwidth]{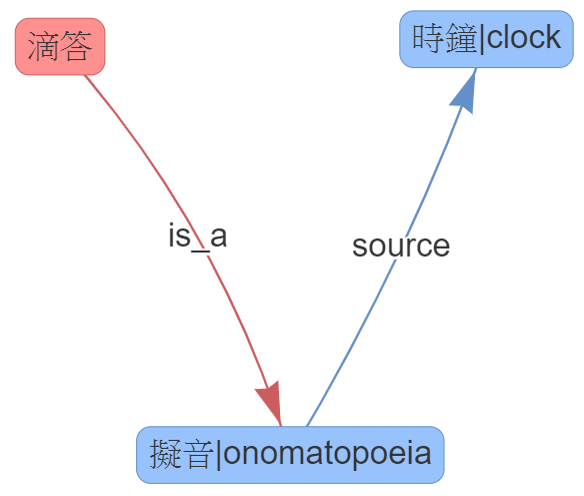}
        \caption{\label{fig:parse:tick-tock}\cc{滴答} (tick-tock).}
    \end{subfigure}
    \hspace{.01\columnwidth}
    \begin{subfigure}{.44\columnwidth}
        \centering
        \includegraphics[width=\columnwidth]{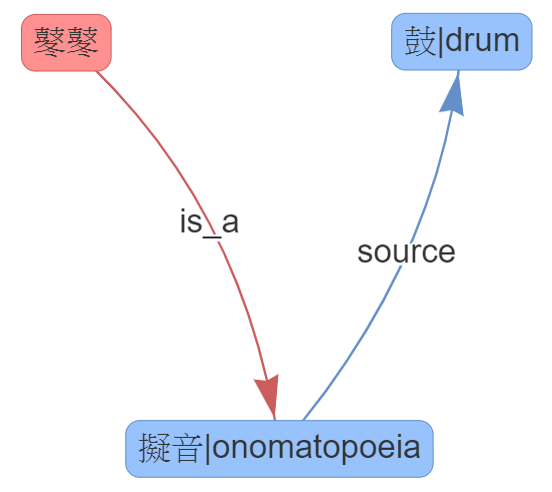}
        \caption{\label{fig:parse:rat-tat}\cc{鼕鼕} (rat-tat).}
    \end{subfigure}
    \hspace{.01\columnwidth}
    \begin{subfigure}{.38\columnwidth}
        \centering
        \includegraphics[width=\columnwidth]{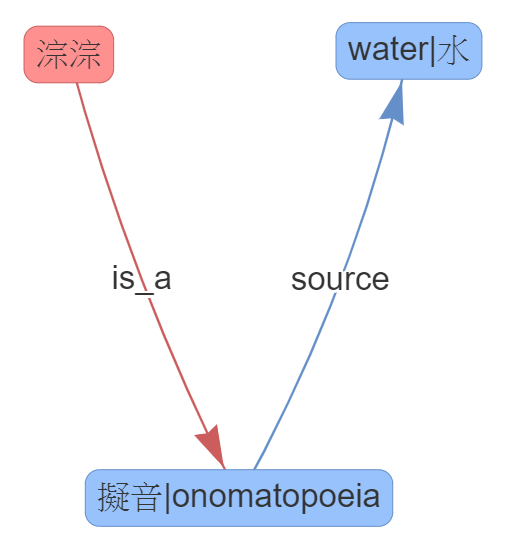}
        \caption{\label{fig:parse:gurgle}\cc{淙淙} (gurgle).}
    \end{subfigure}
    \hspace{.01\columnwidth}
    \caption{Some definition graphs that leads to the sound-origin relation.}
    \label{fig:relation:sound_origin}
\end{figure*}

\begin{figure*}[!t]
    \centering
    \begin{subfigure}{.48\columnwidth}
        \centering
        \includegraphics[width=\columnwidth]{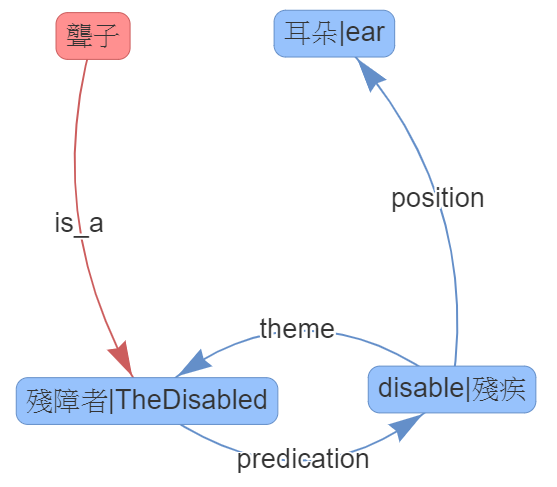}
        \caption{\label{fig:parse:deaf_person}\cc{聾子} (deaf person).}
    \end{subfigure}
    \hspace{.01\columnwidth}
    \begin{subfigure}{.48\columnwidth}
        \centering
        \includegraphics[width=\columnwidth]{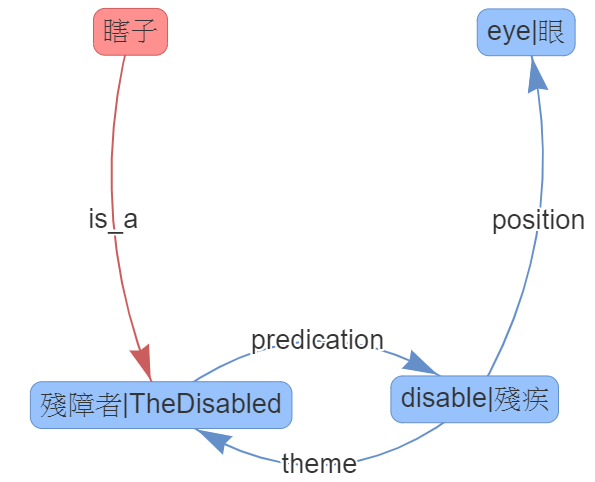}
        \caption{\label{fig:parse:blind_person}\cc{瞎子} (blind person).}
    \end{subfigure}
    \hspace{.01\columnwidth}
    \begin{subfigure}{.48\columnwidth}
        \centering
        \includegraphics[width=\columnwidth]{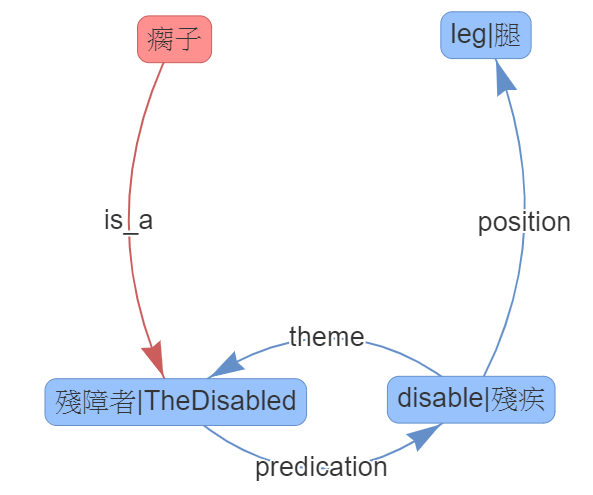}
        \caption{\label{fig:parse:lame_person}\cc{瘸子} (lame person).}
    \end{subfigure}
    \hspace{.01\columnwidth}
    \caption{Some definition graphs that leads to the organ-disabled relation.}
    \label{fig:relation:organ_disabled}
\end{figure*}

\begin{figure*}[!t]
    \centering
    \begin{subfigure}{.44\columnwidth}
        \centering
        \includegraphics[width=\columnwidth]{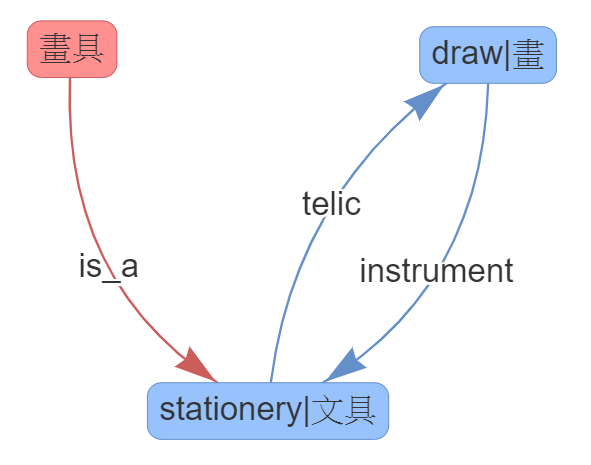}
        \caption{\label{fig:parse:painting_tools}\cc{畫具} (painting tools).}
    \end{subfigure}
    \hspace{.01\columnwidth}
    \begin{subfigure}{.44\columnwidth}
        \centering
        \includegraphics[width=\columnwidth]{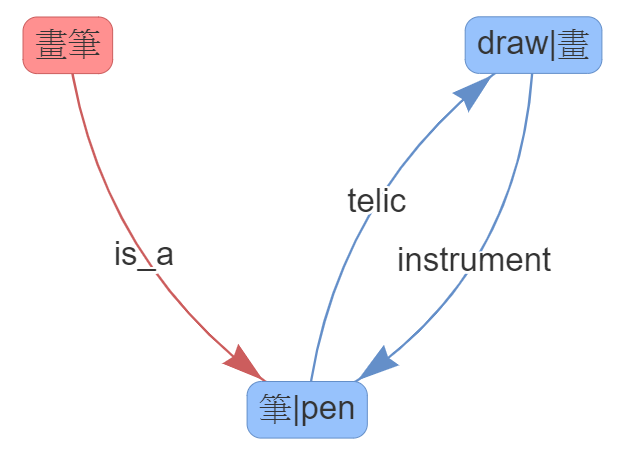}
        \caption{\label{fig:parse:painting_brush}\cc{畫筆} (painting brush).}
    \end{subfigure}
    \hspace{.01\columnwidth}
    \begin{subfigure}{.38\columnwidth}
        \centering
        \includegraphics[width=\columnwidth]{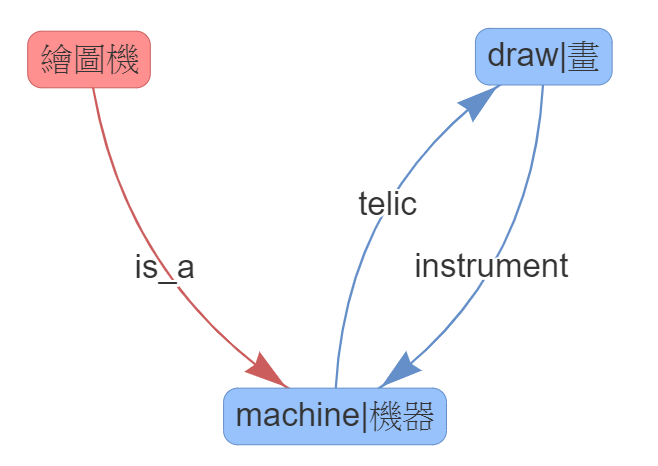}
        \caption{\label{fig:parse:plotter}\cc{繪圖機} (plotter).}
    \end{subfigure}
    \hspace{.01\columnwidth}
    \caption{Some definition graphs that leads to the painter-instrument relation.}
    \label{fig:relation:painter_instrument}
\end{figure*}

\begin{figure*}[!t]
    \centering
    \begin{subfigure}{.48\columnwidth}
        \centering
        \includegraphics[width=\columnwidth]{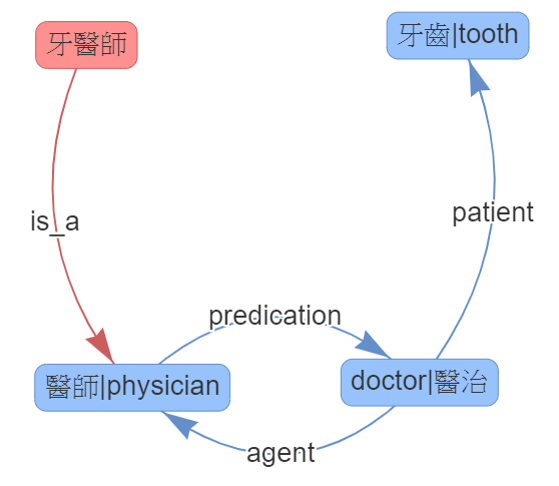}
        \caption{\label{fig:parse:dentist}\cc{牙醫} (dentist).}
    \end{subfigure}
    \hspace{.01\columnwidth}
    \begin{subfigure}{.48\columnwidth}
        \centering
        \includegraphics[width=\columnwidth]{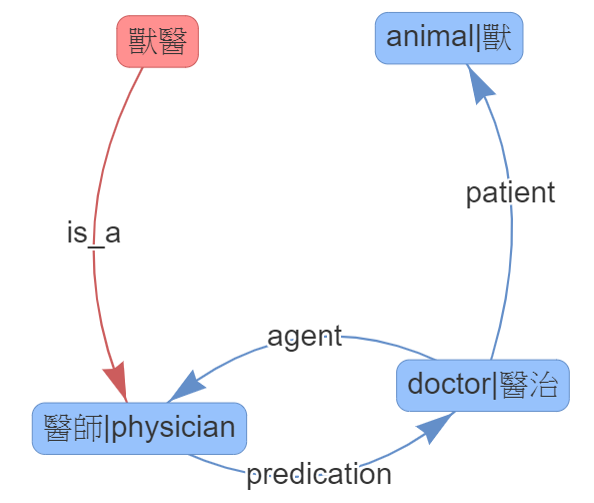}
        \caption{\label{fig:parse:veterinarian}\cc{獸醫} (veterinarian).}
    \end{subfigure}
    \hspace{.01\columnwidth}
    \begin{subfigure}{.48\columnwidth}
        \centering
        \includegraphics[width=\columnwidth]{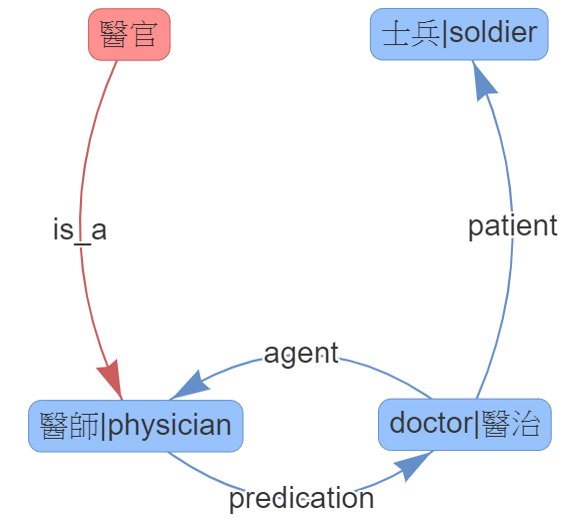}
        \caption{\label{fig:parse:medical_officer}\cc{醫官} (medical officer).}
    \end{subfigure}
    \hspace{.01\columnwidth}
    \caption{Some definition graphs that leads to the doctor-patient relation.}
    \label{fig:relation:doctor_patient}
\end{figure*}

In contrast, all the 90,505 analogies in CA-EHN are automatically extracted from the E-HowNet ontology and then verified by linguists. Still, we can group word pairs into equivalence classes to see what relations are present in the corpus. For example, we have both
\begin{itemize}
    \item ~\cc{樹苗}:\cc{樹}=\cc{蝌蚪}:\cc{青蛙} \\
    (sapling : tree = tadpole : frog)
    \item ~\cc{蝌蚪}:\cc{青蛙}=\cc{孑孓}:\cc{蚊} \\
    (tadpole : frog = wriggler : mosquito)
\end{itemize}
So we can easily know that (\cc{樹苗}, \cc{樹}) and (\cc{孑孓}, \cc{蚊}) belong to the same equivalence class, which seems to express the juvenile-adult relation. By grouping all 90,505 commonsense analogies into equivalence classes, we find that CA-EHN have an unprecedented coverage of 763 relations. Figures~\ref{fig:relation:sound_origin}, \ref{fig:relation:organ_disabled}, \ref{fig:relation:painter_instrument}, \ref{fig:relation:doctor_patient} show some of the relations.

\subsection{Embedding Benchmarking}

To evaluate the robustness of using CA-EHN for the classic intrinsic embedding evaluation task (Section~\ref{sec:evaluation}), we trained and tested different word embeddings across different benchmark datasets.

We trained each word embedding using either GloVe \cite{Pennington:EMNLP2014} or SGNS \cite{Mikolov:NIPS2013} on a small or a large corpus. The small corpus consisted of the traditional Chinese part of Chinese Gigaword \citelanguageresource{Graff:LDC2003} and ASBC 4.0 \citelanguageresource{Ma:ROCLING2001}. The large corpus additionally included the Chinese part of Wikipedia. When calculating accuracy, only those analogy questions of which all words were in an embedding were considered. So a smaller dictionary was not penalized by lower analogy question coverage.

Table~\ref{tab:embedding_benchmarking} shows the results of different combinations of embeddings and benchmarks. It can be seen that CA-EHN is a robust benchmark for the analogy task. On all existing benchmarks and CA-EHN, it is consistent that GloVe-Small is the worst-performing and SGNS-Large is the best. Furthermore, the new dedicated commonsense analogy corpus appears substantially more challenging than existing analogies for distributed word representations.

\subsection{Commonsense Benchmarking}
\label{sec:commonsense_benchmarking}

Two central hypotheses of this work are that it is useful to embed more commonsense knowledge and that CA-EHN tests this aspect of word embedding. To verify these hypotheses, we infused some structure knowledge of commonsense ontology to word embeddings and observed their performance change across benchmarks.

We infused distributed word representations with the hypo-hyper and same-taxon knowledge in the E-HowNet taxonomy (Section~\ref{sec:taxonomy}) and the HIT-Thesaurus\footnote{\cc{同義詞詞林擴展版}(\url{https://github.com/taozhijiang/chinese_correct_wsd})} through retrofitting \cite{Faruqui:NAACL2015}. For example, in Figure~\ref{fig:taxonomy}, the word vector of \cc{物體} was optimized to be close to both its distributed representation and the word vectors of \cc{物質} (same-taxon) and \cc{東西} (hypo-hyper).

Table~\ref{tab:ehownet_retrofit_benchmarking},~\ref{tab:hitthesaurus_retrofit_benchmarking} shows the results of different combinations of retrofitted embeddings and benchmarks. Firstly, retrofitted embeddings achieve better performance on most existing datasets, suggesting the benefits of embedding more commonsense knowledge. Secondly, on CA-EHN, each retrofitted embedding significantly outperforms its pure distributed counterpart in Table~\ref{tab:embedding_benchmarking}. Performance increases by up to 179\% and 88\% by infusing E-HowNet taxonomy and HIT-Thesaurus respectively. This shows that CA-EHN is a great indicator of how well word representations embed commonsense knowledge.

\section{Conclusion}

We have presented CA-EHN, a large and dedicated commonsense word analogy dataset, by leveraging word sense definitions in E-HowNet. After linguist checking, we have 90,505 Chinese analogies covering 5,656 words and 763 commonsense relations. The experiments showed that CA-EHN could become an important benchmark for testing how well future embedding methods capture commonsense knowledge, which is crucial for models to generalize inference beyond their training corpora. With translations provided by E-HowNet, Chinese words in CA-EHN can be transferred to English multi-word expressions.

\section{Acknowledgements}

We are grateful for the insightful comments from anonymous reviewers. This work is supported by the Ministry of Science and Technology of Taiwan under grant numbers 109-2634-F-001-010, 109-2634-F-001-008.

\section{Bibliographical References}
\label{main:ref}

\bibliographystyle{lrec}
\bibliography{lrec2020W-xample-kc}

\section{Language Resource References}
\bibliographystylelanguageresource{lrec}
\bibliographylanguageresource{languageresource}

\end{document}